\documentclass[12pt]{article}

\usepackage{sbc-template}

\usepackage{graphicx,url}
\usepackage{acronym}
\usepackage{multirow}
\usepackage{booktabs}
\usepackage[utf8]{inputenc}  
\usepackage{subfigure}
\usepackage{float}
\usepackage{xcolor}

\usepackage[absolute]{textpos}
\title{Improving Sickle Cell Disease Classification:\\ A Fusion of Conventional Classifiers, Segmented Images, \\ and Convolutional Neural Networks}

\author{Victor Júnio Alcântara Cardoso\inst{1}, Rodrigo Moreira\inst{1}, \\João Fernando Mari\inst{1}, Larissa Ferreira {Rodrigues Moreira}\inst{1}\inst{2}}

\address{Institute of Exacts and Technological Sciences -- Federal University of Viçosa (UFV)\\
  Rio Paranaíba -- MG -- Brazil
\nextinstitute School of Computer Science -- Federal University of Uberlândia (UFU)\\
  Uberlândia -- MG -- Brazil
  \email{\{victor.cardoso, rodrigo, joaof.mari, larissa.f.rodrigues\}@ufv.br}
  \email{larissarodrigues@ufu.br}
}

\begin{document} 

%-----A--------
\acrodef{AI}{Artificial Intelligence}
\acrodef{AIaaS}{Artificial Intelligence as a Service}
%-----C--------
\acrodef{CNNs}{Convolutional Neural Networks}

%-----K--------
\acrodef{KNN}{K-Nearest Neighbors}

%-----M--------
\acrodef{MLP}{Multi Layer Perceptrons}

%-----S--------
\acrodef{SVM}{Support Vector Machines}
\acrodef{SGD}{Stochastic Gradient Descent}
\acrodef{SIFT}{Scale Invariant Feature Transform}
\acrodef{SURF}{Speeded-Up Robust Features}

\begin{textblock*}{15cm}(3cm,1.3cm) % (Largura, (x,y))
\noindent
    \footnotesize\textcolor{red}{This paper has been accepted by the Encontro Nacional de Inteligência Artificial e Computacional (ENIAC) 2023. The definite version of this work was published by SBC-OpenLib as part of the ENIAC conference proceedings. 
    DOI: \url{https://doi.org/10.5753/eniac.2023.234076}}
\end{textblock*}

\maketitle

\begin{abstract}
Sickle cell anemia, which is characterized by abnormal erythrocyte morphology, can be detected using microscopic images. Computational techniques in medicine enhance the diagnosis and treatment efficiency. However, many computational techniques, particularly those based on Convolutional Neural Networks (CNNs), require high resources and time for training, highlighting the research opportunities in methods with low computational overhead. In this paper, we propose a novel approach combining conventional classifiers, segmented images, and CNNs for the automated classification of sickle cell disease. We evaluated the impact of segmented images on classification, providing insight into deep learning integration. Our results demonstrate that using segmented images and CNN features with an SVM achieves an accuracy of 96.80\%. This finding is relevant for computationally efficient scenarios, paving the way for future research and advancements in medical-image analysis.

\end{abstract}
     
%\begin{resumo} 
 % Este meta-artigo descreve o estilo a ser usado na confecção de artigos e
  %resumos de artigos para publicação nos anais das conferências organizadas
  %pela SBC. É solicitada a escrita de resumo e abstract apenas para os artigos
  %escritos em português. Artigos em inglês deverão apresentar apenas abstract.
  %Nos dois casos, o autor deve tomar cuidado para que o resumo (e o abstract)
  %não ultrapassem 10 linhas cada, sendo que ambos devem estar na primeira
  %página do artigo.
%\end{resumo}

\section{Introduction}

Sickle cell anemia is a hereditary disease that results in the irregular shaping of erythrocytes, causing blockages in blood flow throughout the body. Instead of the smooth, circular shape characteristic of normal cells, individuals with sickle cell anemia have erythrocytes that are sickle-like in shape. This abnormal shape leads to a shortened lifespan of erythrocytes, resulting in anemia. Patients experience various symptoms, including pain crises due to blood vessel obstruction, fatigue, myocardial infarction, cerebrovascular accident (CVA), renal disease, pulmonary embolism, and infections. Early diagnosis, combined with appropriate interventions, significantly reduces mortality in the first five years of life from 25\% to less than 3\%~\cite{hoffman2013hematology}~\cite{sicklecell}~\cite{Kavanagh2022}.

The newborn bloodspot screening test is a valuable tool for early diagnosis of sickle cell disease~\cite{Kavanagh2022}. However, the availability of this diagnostic method is limited in various emerging regions due to constraints such as insufficient financial resources, lack of laboratory supplies, and a shortage of specialized professionals. For instance, in Brazil, 100\% of hospitals in Minas Gerais conduct newborn bloodspot screening tests, while in Amapá, the coverage is only around 55\%~\cite{sicklecell}. 

Image processing and machine learning techniques have played a crucial role in disease detection and have emerged as cost-effective solutions, particularly beneficial for emerging countries and remote regions \cite{Schwalbe2020}. In this context, the automatic classification of sickle cell disease using microscopy images has gained significant attention as a promising and globally accessible alternative.

Conventional classification algorithms, also known as traditional machine learning methods, are commonly utilized in pattern recognition tasks and have found applications in various medical domains~\cite{lucas, Backes2022}. However, these algorithms rely on manual feature extraction. To overcome this limitation, \ac{CNNs} have shown remarkable results compared to traditional classifiers. The key advantage of CNNs is their ability to extract features during training, eliminating the need to preprocess the image~\cite{Goodfellow2016, Ponti2017}.

In this paper, we propose to evaluate the performance of CNNs by combining them with traditional classifiers to support sickle-cell disease diagnosis using microscopy images. The main contribution of this paper relies on the direct mastery of CNN for feature extraction and feeding a classical classifier such as \ac{SVM} and Bayes with these features, requiring lower time to train and predict while demanding low resources footprint. Additionally, our approach is suitable for retrieving relevant features from images using an \ac{AIaaS} Architecture~\cite{RodriguesMoreira2023}.

The remaining of this paper is organized as follows: Section~\ref{sec:related-work} provides an overview of the related work conducted in the field. Section~\ref{sec:material_methods} details the proposed approach. The results obtained from the experiments are presented and discussed in Section~\ref{sec:results}. Finally, we conclude the paper in Section~\ref{sec:conclusion} and discuss potential avenues for future research.

\section{Related Work}\label{sec:related-work} 

Recent advancements in \ac{AI} have significantly impacted medical imaging diagnostics across various domains, such as radiology, histology, oncology, and dermatology~\cite{Carmo2021, Backes2022, RodriguesMoreira2023}. These advancements have improved diagnostic accuracy and efficiency, leading to the development of robust computational methods~\cite{Zhou2021}. In this section, we provide an overview of studies that have explored the utilization of \ac{AI} methods for diagnosing sickle cell disease.

The classification of sickle cell disease in the \textit{erythrocytesIDB} dataset was initially proposed by \cite{gual2015erythrocyte}. They employed an active contour segmentation technique based on integral geometry to extract features from erythrocytes. The erythrocytes classification was performed using the \ac{KNN} algorithm.

\cite{rodrigues2016} presented an approach that focused on image segmentation to distinguish normal, sickle, and other deformed cells from the background in microscopy images. They applied morphological feature extraction, considering a set of nine simple geometric shape features. To identify the most discriminative features, they employed univariate ANOVA. The study evaluated three classifiers: Naive Bayes, \ac{SVM}, and \ac{KNN}. The results indicated that the \ac{SVM} classifier achieved the highest performance using the selected feature set, excluding the perimeter.

\cite{lucas} introduced a novel feature encoding process for red blood cell classification. They combined \ac{SIFT} and \ac{SURF} key point descriptors by stacking them into a single matrix. The authors then evaluated the performance of two classification algorithms, \ac{SVM} and \ac{MLP}, in classifying the red blood cells. The experimental results demonstrated that the SVM classifier achieved the best performance. 

\cite{Silva2020} proposed a method that leveraged data augmentation techniques in conjunction with a CNN to enhance sickle cell disease classification performance. They employed the Bayesian optimization algorithm to identify an effective data augmentation strategy specifically tailored for erythrocyte images. Moreover, they utilized a lightweight CNN architecture for the classification task. 

\cite{Petrovic2020} introduced an approach to effectively analyze red blood cell morphology by selecting the optimal classification method and features. Their study evaluated seven conventional classifiers, including \ac{SVM}, and an optimization strategy to identify the best-performing approach. For feature extraction, the authors leveraged established methods from the literature, encompassing shape and texture features. Machine learning techniques were then employed for the task of morphology classification.

\cite{Paz-Soto2020} presented an approach to classify sickle cell disease in microscopy images by employing neural networks trained with features extracted using integral geometry-based functions. By utilizing these functions, which capture contour information of red blood cells, the authors aimed to improve classification accuracy. However, one potential drawback of using integral geometry-based features is their limited descriptive power for capturing complex structural variations in red blood cells and sensibility to noise or small perturbations in the image, potentially leading to misclassification or reduced accuracy in some instances. 

The study conducted by \cite{Alzubaidi2020} utilized transfer learning to distinguish sickle cells from normal red blood cells across three distinct datasets: a main dataset, a transfer learning training dataset, and a testing dataset. The authors proposed three CNN architectures and achieved accuracy levels exceeding 98\% by leveraging transfer learning, feature extraction, and SVM as the classifier. However, it should be noted that training a CNN with a dataset from the same domain is often infeasible in real-world scenarios.

Among the various studies reviewed, the majority have focused on the direct utilization of morphological features extracted from the contour of red blood cells. These features encompass circular and elliptical shape coefficients, roundness, eccentricity, perimeter, area, diameter, and integral geometry, among others. Notably, using shape features in conjunction with classifiers such as \ac{SVM} and Neural Networks has shown promising performance.

%Despite the various works reviewed, most studies have focused on utilizing morphological features directly extracted from the contour of red blood cells, such as circular and elliptical shape coefficients, roundness, eccentricity, perimeter, area, diameter, integral geometry, among others. Notably, using shape features in conjunction with classifiers such as \ac{SVM} and Neural Networks has shown promising performance.

However, these previous approaches are predominantly based on manual feature design and selection, which can introduce subjectivity and consume a significant amount of time. Additionally, this process necessitates specialized knowledge and may impede the scalability and generalizability of the approach to different datasets. In contrast, deep learning-based classification mitigates the need for extensive preprocessing techniques. Nevertheless, the computational cost of training deep learning models can pose limitations when deploying them in real-world scenarios.

To the best of our knowledge, there is a lack of studies that utilize segmented images to classify red blood cells using the combination of conventional classifiers with deep learning techniques. Therefore, our study aims to enhance the classification of sickle cell disease by integrating traditional classifiers, segmented images, and \ac{CNNs}. We also investigate the performance of our proposed approach on both original and segmented images to determine the most effective method for aiding in the diagnosis of sickle cell disease.

\section{Proposed Approach}\label{sec:material_methods} 

\begin{figure}[!htbp]
    \centering
    \includegraphics[width=\linewidth]{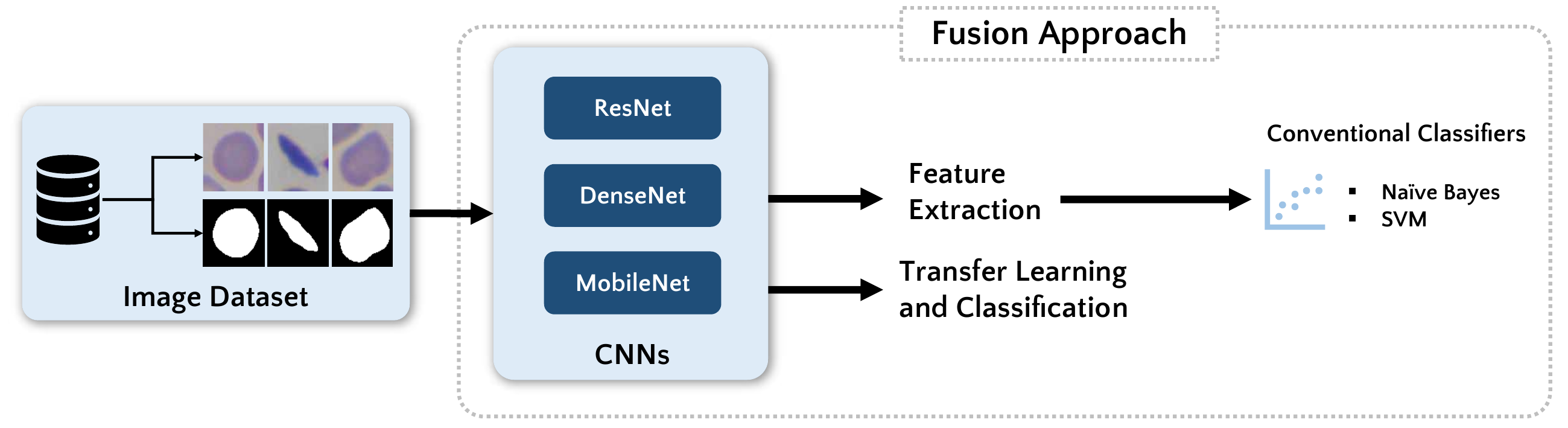}
    \caption{Steps of the proposed approach.}
    \label{fig:steps}
\end{figure}

% This section presents our proposed approach for improving the classification of sickle cell disease. Our main goal is to develop a comprehensive and effective method that accurately classifies sickle cell disease using preprocessed images and a fusion of traditional and deep learning techniques. Our study specifically focuses on identifying the best classification strategy to enhance overall performance. The steps of our proposed method are depicted in Figure~\ref{fig:steps}.
This section presents our proposed approach for improving the classification of sickle cell disease. Our main goal is to develop a comprehensive and effective method that accurately classifies sickle cell disease using a fusion of traditional and deep learning techniques. We also compared feeding the \ac{CNNs} with the original and segmented images. Our study specifically focuses on identifying the best classification strategy to enhance overall performance. The steps of our proposed method are depicted in Figure~\ref{fig:steps}.

\subsection{Dataset}
\begin{figure}[!htbp]
    \centering
    \includegraphics[width=0.55\linewidth]{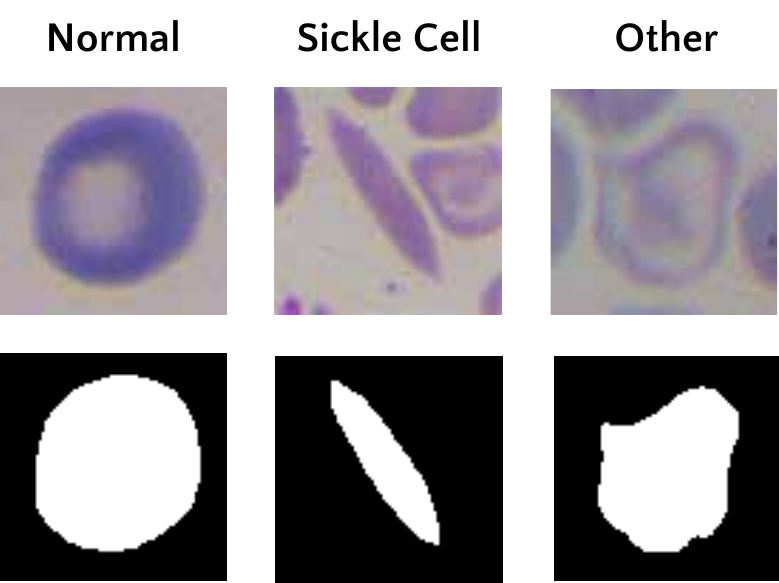}
    \caption{Image instances from the \textit{erythrocytesIDB} dataset showing original (top) and segmented (bottom) images.}
    \label{fig:dataset}
\end{figure}

The images used in this study were obtained from the \textit{erythrocytesIDB} dataset \footnote{Available in: \url{http://erythrocytesidb.uib.es/}} provided by the University of the Balearic Islands~\cite{gonzalez2014red}. The dataset comprises 626 images, each featuring a single, centrally positioned cell, categorized into three classes: normal erythrocytes (202 images), sickle cells (211 images), and erythrocytes with other deformations (213 images). All images are in JPG format and have a resolution of 80 $\times$ 80 pixels. Additionally, we utilized the segmented images generated using the method proposed by \cite{rodrigues2016}. Figure~\ref{fig:dataset} shows some images from each class in the dataset, presenting both the original and segmented versions.

 \bigskip \bigskip
 
\section{Traditional Classifiers}

The classification task is crucial in various fields, including medical imaging, where an accurate and efficient diagnosis is essential~\cite{Zhou2021, Nazir2023}. Traditional classifiers have been widely used for their interpretability and ability to handle high-dimensional data~\cite{Aljuaid2022}. In this study, we utilized two well-established classifiers: \ac{SVM} and Naive Bayes.

The \ac{SVM} is a widely used traditional classifier in machine learning due to its ability to handle complex datasets and its robustness against overfitting. It is particularly effective in handling high-dimensional data and finding optimal decision boundaries. \ac{SVM} works by mapping input data into a higher-dimensional feature space and identifying a hyperplane that maximally separates different classes. It achieves this by maximizing the margin between the classes, allowing for better generalization to unseen data~\cite{Cortes1995}. 

Naive Bayes is another popular traditional classifier, particularly known for its simplicity and efficiency~\cite{Wickramasinghe2021}. It is based on Bayes' theorem and assumes independence among the features. Despite this naive assumption, Naive Bayes often delivers competitive performance. Naive Bayes models estimate the probability of each class given the input features and then classify the data based on the highest probability. The classifier calculates the conditional probabilities using the training data and utilizes these probabilities to make predictions on unseen data. Additionally, Naive Bayes is known for its fast training and prediction times, making it suitable for various applications~\cite{Duda2000}.

We selected these algorithms as our classification models due to their well-established effectiveness and their suitability for our research objective of improving sickle cell disease classification. By employing these classifiers, we aim to investigate their capabilities in accurately classifying the extracted features from the images and compare their performance against other advanced techniques in our study.
%We selected these algorithms as our classification models based on their well-established effectiveness and suitability for our research objectives. By employing these classifiers, we aim to investigate their capabilities in accurately classifying the extracted features from the images and compare their performance against other advanced techniques in our study.

\section{Evaluated Architectures}
Convolutional Neural Networks (CNNs) are a powerful class of deep learning models widely used in computer vision tasks. They are specifically designed to process and analyze visual data, such as images. By leveraging hierarchical feature extraction and parameter sharing, CNNs can learn complex patterns and structures in images~\cite{Goodfellow2016}\cite{Ponti2017}. In this study, we investigated the effectiveness of three \ac{CNNs}: DenseNet-169, ResNet-50, and MobileNet, chosen due to their success in previous image classification tasks.

DenseNet architecture has demonstrated excellent performance in image classification tasks~\cite{Zhou2022}. It introduces the concept of dense connections, where each layer is directly connected to every other layer in a feed-forward fashion. This connectivity pattern allows for better information flow and feature reuse, improving gradient propagation and alleviating the vanishing gradient problem~\cite{Huang2016}. We considered in this study the DenseNet-169.

ResNet, short for Residual Network, is a popular deep neural network architecture known for its residual connections. These connections enable the network to learn residual mappings, making it easier to train deeper networks. By bypassing certain layers and allowing direct information flow, ResNet can effectively tackle the challenges of training very deep neural networks, leading to improved accuracy and performance~\cite{He2016}. In this study, we evaluated the ResNet with 50 layers.

MobileNet is a lightweight deep neural network architecture designed for efficient computation and deployment on resource-constrained devices. It utilizes depth-wise separable convolutions to reduce computational complexity while maintaining good accuracy. MobileNet achieves a good balance between model size and accuracy, making it suitable for applications with limited computational resources, such as mobile devices and embedded systems~\cite{Howard2017}.

We selected DenseNet-169, ResNet-50, and MobileNet for our study due to their unique structural differences as detailed in~\cite{RodriguesMoreira2023}. Through the evaluation of these networks, our objective is to assess their specific advantages and limitations in the classification of sickle cell disease. This analysis will provide valuable insights into the impact of structural variations on both model performance and feature extraction, enhancing our understanding of the effectiveness of \ac{CNNs} in this domain. We used these architectures as feature extractors by removing the classification layer at the end of training. We extracted 232,736 features from DenseNet-169, 90,947 features from ResNet-50, and 33,792 features from MobileNet.

\section{Validation Protocol}

All CNN models were trained and tested using the stratified $k$-fold cross-validation approach ($k$=5) to assess the classification performance. This involved dividing the dataset into five subsets, using one subset as the validation set while training the classifiers on the remaining subsets. By repeating this process for each subset, we obtained a comprehensive evaluation of the classifiers' performance, minimizing bias and variability in the results.

We present our findings by calculating the average accuracy, precision, recall, and F1-score~\cite{Duda2000}, which are aggregated across the $5$-fold cross-validation process. 

\begin{itemize}
    \item Accuracy: measures the overall correctness of the classification (Eq.~\ref{eq:acc2}).
    
    \begin{equation}
    \label{eq:acc2}
        Accuracy = \frac{TP + TN}{TP + TN + FP + FN} 
    \end{equation}
    
    \item Precision: quantifies the proportion of correctly classified positive instances (Eq.~\ref{eq:prc}).
    
    \begin{equation}
    \label{eq:prc}
        Precision = \frac{TP}{TP + FP}  
    \end{equation}

    \item Recall: determines the ability to identify all positive instances (Eq.~\ref{eq:rec}).

    \begin{equation}
    \label{eq:rec}
        Recall = \frac{TP}{TP + FN}  
    \end{equation}

    \item F1-Score: provides a harmonic mean of precision and recall, offering a balanced assessment of the classifier's performance (Eq.~\ref{eq:fsc}).
 
    \begin{equation}
    \label{eq:fsc}
        F1\!\!-\!Score  = 2 \times \frac{Precision \times Recall}{Precision + Recall}
    \end{equation}   
    
\end{itemize}\smallskip

Where $TP$ represents true positives, $TN$ denotes true negatives, $FP$ signifies false positives, and $FN$ indicates false negatives. For each fold, we calculate these metrics and subsequently calculate the average metrics across all folds.

\section{Results and Discussion} \label{sec:results} 

All experiments were conducted using Python (version 3.6) and implemented in Keras~\cite{Chollet2015keras} with TensorFlow as the backend. The classification evaluation was performed on Google Colaboratory, utilizing an Intel(R) Xeon(R) 2.30GHz processor, 12GB RAM, and the NVIDIA Tesla T4 GPU. Aiming at reproducibility, our code is available and the results in the open-source repository\footnote{Source code available on: \\ \footnotesize{\url{https://github.com/larissafrodrigues/sickle-cell-classification-ENIAC2023}}}.

\textbf{CNNs as Feature Extractors.} Initially, we conducted experiments to assess the performance of CNNs as feature extractors. Initially, the original images were used without any preprocessing. The features extracted by the CNNs were then employed as feature vectors for conventional classifiers, namely Naive Bayes and linear \ac{SVM}. For the linear \ac{SVM}, a constant of $2.9$ was empirically defined.

Table \ref{tab:classconv-originais} presents the performance of each classifier, with the best result achieved by extracting features using the MobileNet architecture and classifying with SVM, achieving an accuracy of 91.21\%. When considering feature extraction with ResNet-50, it is evident that both evaluated classifiers did not achieve satisfactory results.

\begin{table}[!htbp]
\renewcommand{\arraystretch}{1.3}
\centering
\caption{Classification performance considering CNN as feature extractor and original images.}
\label{tab:classconv-originais}
\begin{tabular}{lcccccccc}
\cmidrule{2-9}
\multicolumn{1}{c}{\textbf{}} & \multicolumn{2}{c}{\textbf{Accuracy (\%)}}        & \multicolumn{2}{c}{\textbf{Precision (\%)}}        & \multicolumn{2}{c}{\textbf{Recall (\%)}}          & \multicolumn{2}{c}{\textbf{F1-Score (\%)}}        \\ \cmidrule{2-9}
\multicolumn{1}{c}{\textbf{}} & \textbf{Bayes} & \multicolumn{1}{l}{\textbf{SVM}} & \textbf{Bayes} & \multicolumn{1}{l}{\textbf{SVM}} & \textbf{Bayes} & \multicolumn{1}{l}{\textbf{SVM}} & \textbf{Bayes} & \multicolumn{1}{l}{\textbf{SVM}} \\ \hline
\textbf{DenseNet}         & 82.60          & 90.59                            & 83.47          & 91.23                            & 82.67          & 90.66                            & 82.82          & 90.61                            \\
\textbf{ResNet}            & 59.10          & 68.05                            & 61.30          & 70.26                            & 59.73          & 67.88                            & 52.91          & 64.69                            \\
\textbf{MobileNet}            & 85.30          & \textbf{91.21}                   & 87.48          & \textbf{91.42}                   & 85.30          & \textbf{91.28}                   & 85.39          & \textbf{91.07} \\ \hline                 
\end{tabular}
\end{table}

Subsequently, experiments were conducted to assess the impact of segmenting images using the method proposed by \cite{rodrigues2016} on feature extraction. The experimental results, shown in Table \ref{tab:classconv-seg}, demonstrate that segmented images improved the performance of both evaluated classifiers. It is evident that when classifying segmented images, for all \ac{CNNs} used as feature extractors, the Naive Bayes classifier achieved an accuracy higher than 90\%, and the \ac{SVM} classifier achieved an accuracy higher than 95\%.

\begin{table}[!htbp]
\renewcommand{\arraystretch}{1.3}
\centering
\caption{Classification performance considering CNN as feature extractor and segmented images.}
\label{tab:classconv-seg}
\begin{tabular}{lcccccccc} 
\cmidrule{2-9}
\multicolumn{1}{c}{\textbf{}} & \multicolumn{2}{c}{\textbf{Accuracy (\%)}}        & \multicolumn{2}{c}{\textbf{Precision (\%)}}        & \multicolumn{2}{c}{\textbf{Recall (\%)}}          & \multicolumn{2}{c}{\textbf{F1-Score (\%)}}        \\ \cmidrule{2-9}
\multicolumn{1}{c}{\textbf{}} & \textbf{Bayes} & \multicolumn{1}{l}{\textbf{SVM}} & \textbf{Bayes} & \multicolumn{1}{l}{\textbf{SVM}} & \textbf{Bayes} & \multicolumn{1}{l}{\textbf{SVM}} & \textbf{Bayes} & \multicolumn{1}{l}{\textbf{SVM}} \\ \hline
\textbf{DenseNet}         & 95.37          & 95.84                            & 95.74          & 96.00                            & 95.38          & 95.87                            & 95.44          & 95.88                            \\
\textbf{ResNet}            & 90.26          & 96.32                            & 90.98          & 96.55                            & 90.26          & 96.36                            & 90.41          & 96.33                            \\
\textbf{MobileNet}            & 94.26          & \textbf{96.80}                   & 94.69          & \textbf{96.88}                   & 94.26          & \textbf{96.84}                   & 94.31          & \textbf{96.81} \\ \hline                  
\end{tabular}
\end{table}

When considering classification with SVM and feature extraction using ResNet-50, the results increased from 68.05\% when using original images (Table \ref{tab:classconv-originais}) to 96.32\% when using segmented images (Table \ref{tab:classconv-seg}). This result suggests that ResNet-50 was able to extract important shape patterns. Finally, the best result was achieved when considering feature extraction with MobileNet and segmented images, achieving an accuracy of 96.80\%.
\bigskip

% \textbf{CNN with Transfer Learning.} We conducted experiments to evaluate the performance of \ac{CNNs} as classifiers, where the features are extracted by the network during training, and classification is performed afterward. Once again, we evaluated the performance of each CNN when trained with original and segmented images, which were resized to 224 $\times$ 224 pixels (the input size allowed by each CNN). The training process was conducted by fine-tuning a CNN pre-trained with ImageNet~\cite{krizhevsky2012imagenet} over 50 epochs, with a learning rate of 1$\times10^{-4}$, optimization using \ac{SGD}, a momentum coefficient of $0.9$, and data augmentation based on horizontal and vertical flips.

\textbf{CNN with Transfer Learning.} We conducted experiments to evaluate the performance of \ac{CNNs} as classifiers. We also used the trained \ac{CNNs} as feature extractors by removing the top layer, and using the features to feed traditional classifiers. Once again, we evaluated the performance of each CNN when trained with original and segmented images, which were resized to 224 $\times$ 224 pixels (the input size allowed by each CNN). The training process was conducted by fine-tuning a CNN pre-trained with ImageNet~\cite{krizhevsky2012imagenet} over 50 epochs, with a learning rate of 1$\times10^{-4}$, optimization using \ac{SGD}, a momentum coefficient of $0.9$, and data augmentation based on horizontal and vertical flips.

Tables \ref{tab:Originais_TL} and \ref{tab:Seg_TL} present the results obtained when classifying the original and segmented images, respectively. It is evident that the use of segmented images brought improvements to the ResNet-50 and MobileNet architectures. In particular, the results obtained by MobileNet demonstrate a significant improvement in all classification metrics when using segmented images compared to original images. The accuracy increased from 87.70\% with original images to 95.54\% with segmented images. 

\begin{table}[!htbp]
\renewcommand{\arraystretch}{1.3}
\centering
\caption{Classification performance for each CNN considering original images.}
\label{tab:Originais_TL}
\begin{tabular}{lcccc} \hline
\multicolumn{1}{c}{\textbf{CNN}} & \textbf{Accuracy (\%)} & \textbf{Precision (\%)} & \textbf{Recall (\%)} & \textbf{F1-Score (\%)} \\ \hline
\textbf{DenseNet}                     & \textbf{96.65}                 & \textbf{96.89}                  & \textbf{96.71 }               & \textbf{96.62}                  \\
ResNet                       & 92.65                  & 93.13                  & 92.70                & 92.56                  \\
MobileNet                        & 87.70                  & 88.73                  & 87.85                & 87.11  \\ \hline               
\end{tabular}
\end{table}

\begin{table}[!htbp]
\renewcommand{\arraystretch}{1.3}
\centering
\caption{Classification performance for each CNN considering segmented images.}
\label{tab:Seg_TL}
\begin{tabular}{lcccc} \hline
\multicolumn{1}{c}{\textbf{CNN}} & \textbf{Accuracy (\%)} & \textbf{Precision (\%)} & \textbf{Recall (\%)} & \textbf{F1-Score (\%)} \\ \hline
\textbf{DenseNet}                     & \textbf{96.65}                 & \textbf{96.78}                  & \textbf{96.67}               & \textbf{96.65}                  \\
ResNet                        & 95.37                  & 93.50                  & 95.42                & 95.38                  \\
MobileNet                        & 95.54                  & 95.62                  & 95.59                & 95.54  \\ \hline               
\end{tabular}
\end{table} 

This improvement suggests that the segmentation process proposed by \cite{rodrigues2016} effectively enhances the discriminative features captured by MobileNet, resulting in more accurate classifications. The deep convolutional layers of MobileNet enable it to effectively leverage the spatial information provided by the segmentation, allowing it to capture intricate patterns and structures in the images. These findings highlight the potential benefits of utilizing preprocessed images in certain scenarios, as it can significantly improve the performance of CNN-based image classification models.

On the other hand, the experimental results demonstrate that the DenseNet architecture achieved the same result for both evaluations, whether using original images or segmented images. This finding suggests that DenseNet is capable of effectively capturing important features from the original images alone, without the need for additional preprocessing, such as segmentation. %The dense connectivity and feature reuse mechanisms in DenseNet enable it to effectively learn and exploit the inherent patterns and structures present in the original images. 
This result highlights the robustness of DenseNet in handling different types of cell images, making it an alternative for sickle cell image classification tasks where preprocessing steps may not be necessary or yield significant improvements.

\begin{figure}[!htbp]
  \centering
  \subfigure[Original Images]{
    \includegraphics[width=0.462\textwidth]{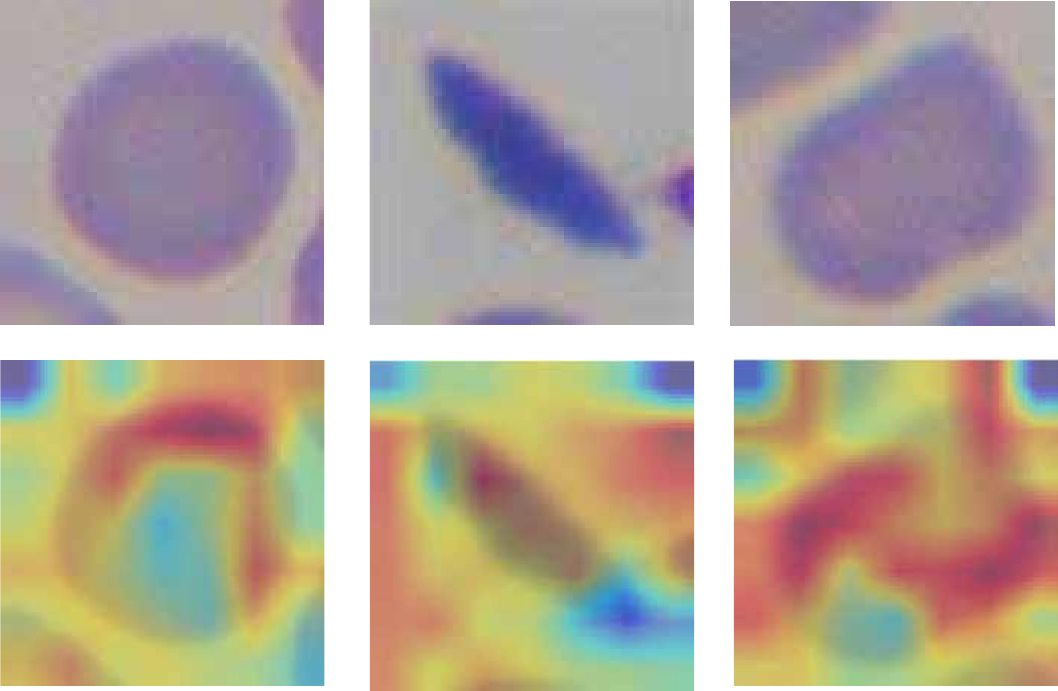}
    \label{fig:map1}
  }
  \hspace{0.5cm}
  \subfigure[Segmented Images]{
    \includegraphics[width=0.462\textwidth]{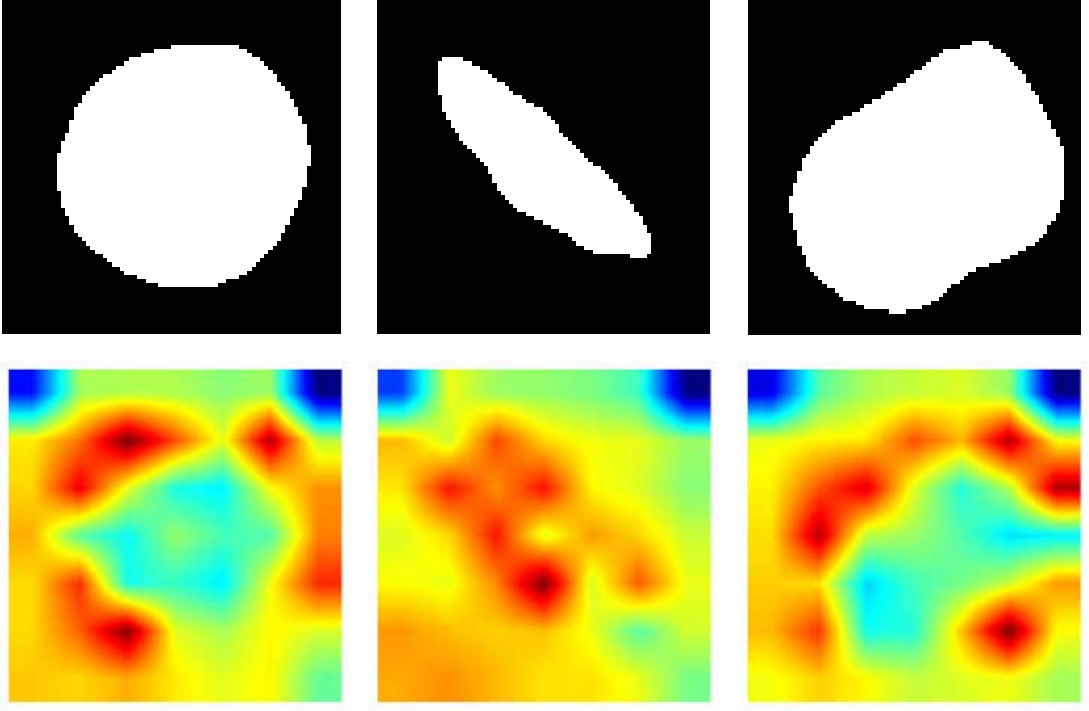}
    \label{fig:map2}
  }
  \caption{Examples of visualization of feature activation obtained by DenseNet.}
  \label{fig:maps}
\end{figure}

Figure~\ref{fig:maps} illustrates an example of feature maps extracted by DenseNet using original and segmented images. By analyzing these maps, it can be observed that DenseNet effectively captures various patterns and textures present in the original images. This demonstrates the ability of DenseNet to learn discriminative representations directly from the raw image data without the need for additional preprocessing steps. 

\begin{figure}[!htbp]
  \centering
  \subfigure[Original Images]{
    \includegraphics[width=0.7\textwidth]{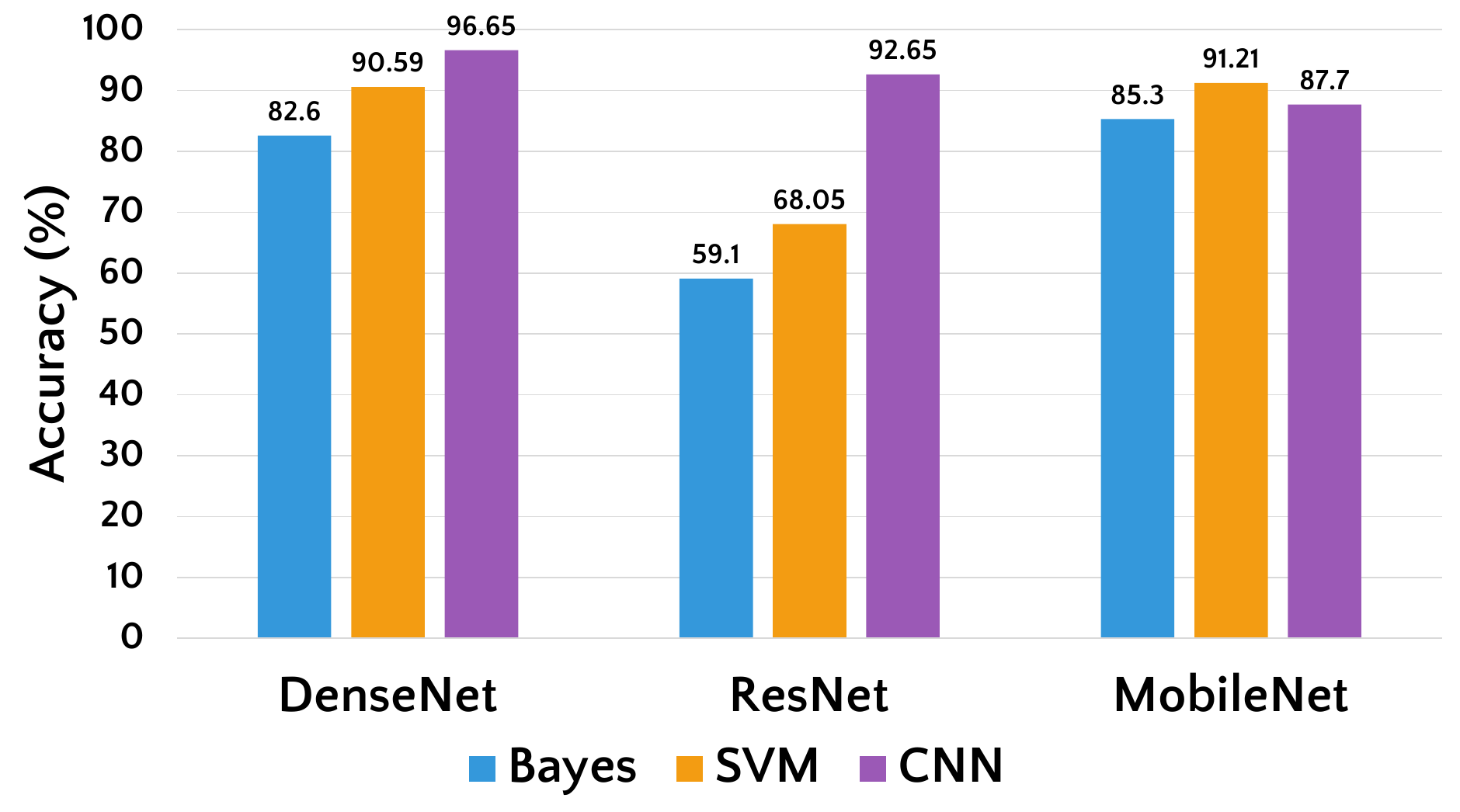}%width=0.462
    \label{fig:subfig1}
  }
  \hspace{0.5cm}
  \subfigure[Segmented Images]{
    \includegraphics[width=0.7\textwidth]{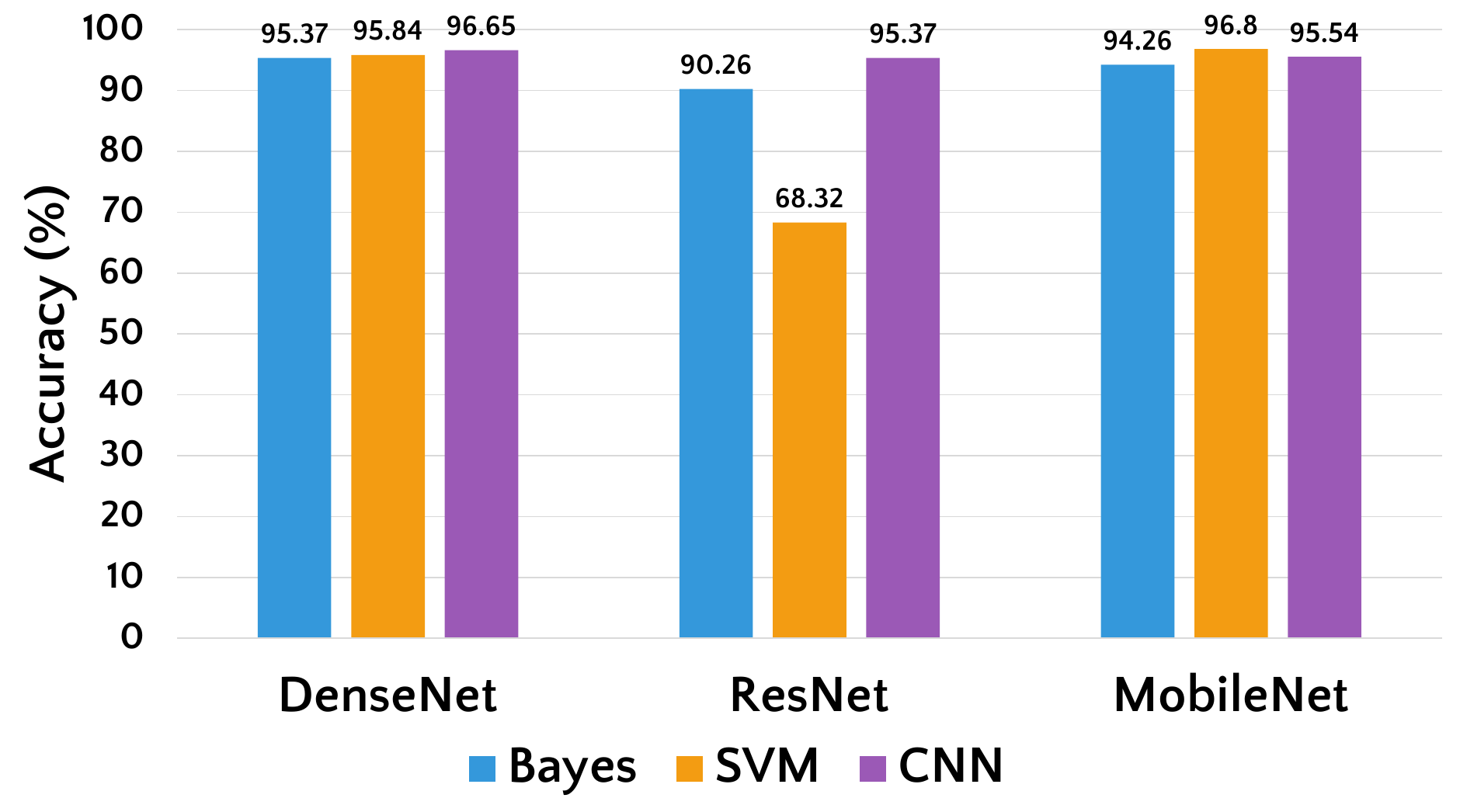}
    \label{fig:subfig2}
  }
  \caption{Comparison between each strategy evaluated considering accuracy performance.}
  \label{fig:charts}
\end{figure}

To summarize the results, the graphs presented in Figure~\ref{fig:charts} illustrate the performance of each CNN architecture in terms of accuracy. The results are shown for two different approaches: classification with the respective CNN itself and feature extraction with \ac{SVM} and Naive Bayes classifiers.

\clearpage
\textbf{Comparison with Literature.} Finally, our best result obtained for the \textit{erythrocytesIDB} dataset was compared with other state-of-the-art approaches in the literature, as shown in Table~\ref{tab:comparacao}. The comparison reveals that our best score is on par with or surpasses the performance of the leading techniques reported in the literature. This comparison underscores the competitiveness of our approach and highlights its effectiveness in addressing the classification of sickle cell disease in microscopy images.

\begin{table}[!htbp]
\renewcommand{\arraystretch}{1.2}
\centering
\caption{Short Comparison with literature.}
\label{tab:comparacao}
\begin{tabular}{lc} \midrule
\multicolumn{1}{c}{\textbf{Approach}}              & \textbf{Accuracy (\%)}  \\ \midrule
\cite{rodrigues2016}                                    & 94.59                   \\
\cite{gual2015erythrocyte}                                   & 96.10                   \\
\cite{lucas}                                   & 93.67                   \\
\cite{Silva2020}                                   & 92.54                   \\
\cite{Paz-Soto2020}                                   & 98.40                   \\
\cite{Alzubaidi2020}                                & 99.98               \\
\textbf{Our approach} & \textbf{96.80}          \\ \midrule
\end{tabular}
\end{table}

\section{Conclusion} \label{sec:conclusion} 

In this paper, we present an innovative method that fuses \ac{CNNs} to extract features with segmented images and traditional classifiers. We validate our approach with images of red blood cells and classify them as healthy, sickle, or with other deformities. After diving into the state of the art we noticed many efforts in using CNN as a classifier neglecting the resource footprint of such approaches leading us to investigate how to improve classical classifiers.

We conducted an investigation into the performance of \ac{CNNs} and conventional classifiers utilizing segmented images in our study. The achieved accuracy of 96.80\% by extracting features with MobileNet and classifying with \ac{SVM} demonstrates that this approach can aid in identifying sickle cell disease in microscopy images.
%Our results suggest that the performance of CNNs and conventional classifiers using segmented images was investigated. The achieved accuracy of 96.80\% by extracting features with MobileNet and classifying with SVM demonstrates that this approach can aid in identifying sickle cell disease in microscopy images.

Although using segmented images requires a preprocessing step, the proposed approach requires less time for the classification task compared to the classification performed by the CNNs themselves while achieving similar accuracies. The proposed approach also offers a runtime approximately 14 times faster.

For future work, we intend to optimize the parameters of \ac{CNNs} to further enhance their performance. Additionally, we will explore the application of other conventional classifiers to evaluate their effectiveness in our classification task. Furthermore, we plan to test the proposed methods on other datasets of medical images and exploit different \ac{CNNs} architectures. These future endeavors will provide a more comprehensive understanding of the capabilities and limitations of different techniques in the field of sickle cell disease classification.

\section*{Acknowledgment}
%Omitted due to the double-blind review.
We would like to thank CAPES and FAPEMIG (Grant number CEX - APQ-02964-17) for the financial support. This study was financed in part by the Coordenação de Aperfeiçoamento de Pessoal de Nível Superior - Brasil (CAPES) - Finance Code 001.

\bibliographystyle{sbc}
\bibliography{references}

\end{document}